\documentclass{article}
\usepackage{ijcai17}

\usepackage{times}
\usepackage{graphicx}
\usepackage{algpseudocode}
\usepackage{bm}
\usepackage{multirow}
\usepackage{tikz-dependency}
\usepackage{url}
\usepackage{subfigure}

\def\argmax{\mathop{\rm argmax}}%
\def\argmin{\mathop{\rm argmin}}%

\title{Joint POS Tagging and Dependency Parsing \\ with Transition-based Neural Networks}
\author{Liner Yang$^1$, Meishan Zhang$^2$, Yang Liu$^1$, Nan Yu$^2$, Maosong Sun$^1$, Guohong Fu$^2$\\
	$^1$State Key Laboratory of Intelligent Technology and Systems  \\
	Tsinghua National Laboratory for Information Science and Technology \\
	Department of Computer Science and Technology, Tsinghua University, Beijing, China \\
	$^2$School of Computer Science and Technology, Heilongjiang University, Harbin, China \\
	\small \{lineryang, mason.zms\}@gmail.com, liuyang2011@tsinghua.edu.cn, yunan.hlju@gmail.com,  sms@tsinghua.edu.cn, ghfu@hlju.edu.cn}

\begin{document}

\maketitle

\begin{abstract}
While part-of-speech (POS) tagging and dependency parsing are observed to be closely related, existing work on joint modeling with manually crafted feature templates suffers from the feature sparsity and incompleteness problems. In this paper, we propose an approach to joint POS tagging and dependency parsing using transition-based neural networks. Three neural network based classifiers are designed to resolve shift/reduce, tagging, and labeling conflicts. Experiments show that our approach significantly outperforms previous methods for joint POS tagging and dependency parsing across a variety of natural languages.
\end{abstract}

\section{Introduction}
% definition of POS tagging and dependency parsing
%Part-of-speech (POS) tagging and dependency parsing are two fundamental tasks for understanding natural languages. While POS tagging aims to assign parts of speech to words in a text to indicate their word categories \cite{Brill:92}, the goal of dependency parsing is to analyze the syntactic structure of sentences by establishing relationships between words \cite{McDonald:05}.
Part-of-speech (POS) tagging \cite{collins:2002,toutanova+:2003,santos:2014,huang+:2015} and dependency parsing \cite{McDonald:05,nivre+:2006,Chen:14,Dyer:15,Kiperwasser:16} are two fundamental tasks for understanding natural languages.
While POS tagging aims to assign parts of speech to words in a text to indicate their word categories, the goal of dependency parsing is to analyze the syntactic structure of sentences by establishing relationships between words.

% POS tagging and dependency parsing can benefit each other
It is widely accepted that POS tagging and dependency parsing are closely related. On one hand, POS tagging often requires long-distance syntactic information for resolving tagging ambiguity \cite{sun+:2013}. Hatori et al. \shortcite{Hatori:11} indicate that the disambiguation between POS tags ``DEG'' (a genitive marker) and ``DEC'' (a complementizer) for a Chinese word {\em de} often depends on global context. On the other hand, as a pre-processing step, POS tagging directly influences the accuracy of dependency parsing significantly. For example, determining the head word of a two-word phrase ``closed door'' directly depends on the POS tag of ``closed'' (adjective or verb in past tense).  Li et al. \shortcite{Li:11} report that dependency accuracy drops by around 6\% on Chinese when automatic POS tagging results instead of ground-truth tags are used.

% Previous work
Therefore, joint POS tagging and dependency parsing has attracted intensive attention in the NLP community. Previous work has focused on jointly modeling POS tagging and dependency parsing using linear models that combine both tagging and parsing features \cite{Li:11,Hatori:11,Bohnet:12,Li:12,zhang+:2012}. Allowing lexicality and syntax to interact in a unified framework, joint POS tagging and dependency parsing improves both tagging and parsing performance over independent modeling significantly \cite{Li:11,Hatori:11,Bohnet:12}.

\begin{figure}[!t]
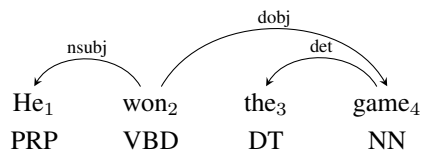

 \begin{minipage}{0.5\textwidth}
      \centering
      \begin{dependency}[arc edge, arc angle=60, text only label, label style={above}]
        \begin{deptext}[column sep=.7cm, row sep=.08cm]
          He$_1$ \& won$_2$ \& the$_3$ \& game$_4$  \\
          PRP \& VBD \& DT \& NN   \\
        \end{deptext}
        %\deproot[edge unit distance=2.5ex]{2}{ROOT}
        \depedge{2}{1}{nsubj}
        \depedge{2}{4}{dobj}
        \depedge{4}{3}{det}
      \end{dependency}
    \end{minipage}
  \vspace{-1.em}
	\caption{Part-of-speech tagging and dependency parsing. Given an English sentence ``He won the game'', our goal is to predict its corresponding part-of-speech tag sequence ``PRP VBD DT NN'' and dependency tree $\{ \langle 2, 1, \textrm{nsubj} \rangle, \langle 4, 3, \textrm{det} \rangle, \langle 2, 4, \textrm{dobj} \rangle \}$.}
	\label{fig:example}
\end{figure}

% Challenge
However, existing work on joint POS tagging and dependency parsing suffers from the feature sparsity and incompleteness problems.  Chen and Manning \shortcite{Chen:14} indicate that lexicalized indicator features indispensable for discriminative dependency parsing are usually highly sparse. The situation in joint POS tagging and dependency parsing is much severer because  tagging and parsing features are concatenated in joint models \cite{Li:11}. Moreover, due to the complexity of tagging and parsing natural languages, it is hard for manually-designed features to cover all regularities. As a result, the incompleteness of feature design is considered as an unavoidable issue in conventional discriminative models \cite{Chen:14}.

% Our work
%In this paper, we propose an approach to joint POS tagging and dependency parsing with translation-based neural networks. Three neural networks based classifiers are designed to resolve shift/reduce, tagging, and labeling conflicts. Experiments show that our approach significantly outperforms previous methods for joint POS tagging and dependency parsing on English Penn Treebank, Chinese Penn Treebank, and Universal Dependencies Treebank across eight natural languages.

In this paper, we propose an approach to joint POS tagging and dependency parsing with neural networks by extending from a transition-based dependency parsing model. Three neural network based classifiers are designed to resolve the conflicts of transition actions, respectively for shift/reduce (dependency parse tree skeletons), tagging (POS tagging), and labeling (dependency label) disambiguations.
%Experiments show that our approach significantly outperforms previous methods for joint POS tagging and dependency parsing on English Penn Treebank (PTB), Chinese Penn Treebank (CTB), and Universal Dependencies Treebanks  (UD) across eight natural languages.
Experiments show that our approach significantly outperforms previous methods for joint POS tagging and dependency parsing on on three treebanks across eight natural languages.

\begin{table*}[!t]
\centering
\begin{tabular}{|l|l|l|}
\hline
{\bf Transition} & {\bf Definition} & {\bf Condition}   \\
\hline \hline
\textproc{Shift} & $\langle S, x_n|B, T, D \rangle \Rightarrow \langle S|x_n, B, T, D \rangle$ & $|B| > 0 \land |T| = N - |B| \land D_{-1}.l \ne \bot$  \\
\textproc{Left} & $\langle S|x_m|x_h, B, T, D \rangle \Rightarrow \langle S|x_h, B, T, D\cup\{ \langle h, m, \bot \rangle \} \rangle$  & $|S| > 1 \land |T| = N - |B| \land D_{-1}.l \ne \bot$ \\
\textproc{Right} & $\langle S|x_h|x_m, B, T, D \rangle \Rightarrow \langle S|x_h, B,T,D\cup\{ \langle h, m, \bot \rangle \} \rangle$ & $|S| > 1 \land |T| = N - |B| \land D_{-1}.l \ne \bot$  \\
\textproc{Tag}$_t$ & $\langle S, B, T, D \rangle \Rightarrow \langle S, B, T\cup \{t \}, D \rangle$ & $|T| = N - |B| - 1$  \\
\textproc{Label}$_l$ & $\langle S|x_h, B, T, D\cup \{ \langle h,m,\bot \rangle \Rightarrow \langle S|x_h, B, T, D\cup \{ \langle h,m,l \rangle \} \rangle$ & $D_{-1}.l = \bot$ \\
\hline
\end{tabular}
\caption{Transitions for joint POS tagging and dependency parsing. We use a quadruple $\langle S, B, T, D \rangle$ to denote a configuration, which consists of a stack $S$, a buffer $B$, a tag sequence $T$, and a dependency arc set $D$. We define five categories of actions \textproc{Shift} (moving a word from the buffer to the stack), \textproc{Left} (generating a right-headed dependency arc), \textproc{Right} (generating a left-headed dependency arc), \textproc{Tag}$_t$ (tagging the last word moved into stack as $t$), and \textproc{Label}$_l$ (labeling the last generated arc as $l$) for the transitions between configurations. We use $\bot$ to denote an undefined syntactic label, and $D_{-1}.l$ to denote the syntactic label of the last generated dependency arc. $N$ is the length of the input sentence.}\label{table:action}
\end{table*}

\section{Approach}

\subsection{Problem Statement}
As shown in Figure \ref{fig:example}, given an English sentence ``He won the game'', the corresponding tag sequence is ``PRP VBD DT NN''. These tags indicate the part of speech of each word: ``He'' is a personal pronoun, ``won'' is a verb in past tense, ``the'' is a determiner, and ``game'' is a noun.

Figure \ref{fig:example} also shows a dependency tree, which is a collection of dependency arcs. The leftmost arc between the first two words indicates that ``won'' is a head word, ``He'' is a modifier, and the syntactic label ``{\em nsubj}'' suggests that ``He'' is a nominal subject.

More formally, given a natural language sentence $\mathbf{x}=x_1, \dots, x_{N}$, we denote its corresponding POS tag sequence as $\mathbf{t}=t_1, \dots, t_N$, where $t \in \mathcal{T}$ is a POS tag and $\mathcal{T}$ is a set of all possible tags. A dependency tree is denoted by $\mathbf{d}=\{ \langle h, m, l \rangle| 0 < h \le N, 0 < m \le N, l \in \mathcal{L} \}$. We use $\langle h, m, l \rangle$ to represent a dependency arc, where $x_h$ is a head word, $x_m$ is a modifier, and $l$ is syntactic label. We use $\mathcal{L}$ to denote the set of all possible syntactic labels. The dependency tree in Figure \ref{fig:example} consists of three arcs: $\langle 2, 1, \textrm{{\em nsubj}} \rangle$, $\langle 2, 4, \textrm{{\em dobj}} \rangle$, and $\langle 4, 3, \textrm{{\em det}} \rangle$.

Therefore, the goal of our work is to generate a tag sequence $\mathbf{t}$ and  a dependency tree $\mathbf{d}$ for a given sentence $\mathbf{x}$.

\subsection{Transition System}
In this work, we leverage a transition-based approach \cite{Nivre:08} to joint POS tagging and dependency parsing, which uses classifiers to predict individual actions of shift-reduce algorithms.

We define a {\em configuration} as  a quadruple $c = \langle S, B, T, D \rangle$, where
\begin{enumerate}
\item $S$: a {\em stack} that is a disjoint sublist of words,
\item $B$: a {\em buffer} that is a sublist of words to be processed,
\item $T$: a {\em tag sequence} that stores the result of POS tagging,
\item $D$: a {\em dependency arc set} that stores the result of dependency parsing.
\end{enumerate}

As shown in Table \ref{table:action}, we define five categories of actions for the transition between configurations: \footnote{While it is possible to integrate two actions into one action (e.g., combining \textproc{Shift} and \textproc{Tag}$_{t}$ into \textproc{Shift}$_t$) \cite{Bohnet:12}, we find that separating tag and label actions (i.e., \textproc{Tag}$_t$ and \textproc{Label}$_l$) from structural actions (i.e., \textproc{Shift}, \textproc{Left}, and \textproc{Right}) leads to significant improvements over using combined actions.}
\begin{enumerate}
\item \textproc{Shift}: move the leftmost word from the buffer $B$ to the stack $S$ ;
\item \textproc{Left}: combine the top two items on the stack, $x_m$ and $x_h$, replace them with $x_h$ as the head, and add an unlabeled dependency arc $\langle h, m, \bot \rangle$ to $D$;
\item \textproc{Right}: combine the top two items on the stack, $x_h$ and $x_m$, replace them with $x_h$ as the head, and add an unlabeled dependency arc $\langle h, m, \bot \rangle$ to $D$;
\item \textproc{Tag}$_t$: assign a POS tag $t$ to the last added word if the previous action is \textproc{Shift} (i.e., $|T|=N-|B|-1$);
\item \textproc{Label}$_l$: assign a syntactic label $l$ to the last generated dependency arc if the previous action is \textproc{Left} or \textproc{Right} (i.e., $D_{-1}.l = \bot$).
\end{enumerate}
where $N$ is the length of the input sentence. We follow Bohnet and Nivre \shortcite{Bohnet:12} to use $\bot$ to denote an undefined syntactic label. $D_{-1}.l$ represents the syntactic label of the last added dependency arc.
Note that the first three actions are used to determine the skeletons of dependency trees,
which can be applied on condition that all words removed from the buffer are tagged (i.e., $|T|=N-|B|$), and all generated dependency arcs are labeled (i.e., $D_{-1}.l \ne \bot$).

Table \ref{table:process} demonstrates the process of joint tagging and dependency parsing for the example in Figure \ref{fig:example}. The initial configuration at step 0 is $c_0 = \langle \emptyset, \{x_1, x_2, x_3, x_4\}, \emptyset, \emptyset \rangle$. In step 1, the action \textproc{Shift} moves the leftmost word $x_1$ (i.e., ``He'') from the buffer $B$ to the stack $S$. Then, the action \textproc{Tag}$_{{\tiny \textrm{PRP}}}$ assigns a POS tag ``PRP'' to the last shifted word ``He''. In this way, the configuration keeps changing by applying various actions until the terminal configuration (i.e., the stack contains only one item, the buffer is empty, all words are tagged, and all arcs are labeled) is generated.
\begin{table*}[!t]
\centering
\begin{tabular}{| c | l | l | l | l | l |}
\hline
Step & Transition & Stack ($S$) & Buffer ($B$) & Tags ($T$) & Dependencies  ($D$) \\
\hline \hline
0 & & & He$_1$  won$_2$  the$_3$  game$_4$ & & \\
1 & \textproc{Shift} & He$_1$ & won$_2$ the$_3$ game$_4$ &  & \\
2 & \textproc{Tag}$_{{\tiny \textrm{PRP}}}$ & He$_1$ & won$_2$ the$_3$ game$_4$ & PRP & \\
3 & \textproc{Shift} & He$_1$  won$_2$ & the$_3$  game$_4$ & PRP & \\
4 & \textproc{Tag}$_{{\tiny \textrm{VBD}}}$ & He$_1$ won$_2$ & the$_3$  game$_4$ & PRP VBD & \\
5 & \textproc{Left} & won$_2$ & the$_3$  game$_4$ & PRP VBD & $\langle 2,1, \bot \rangle$ \\
6 & \textproc{Label}$_{{\tiny \textrm{nsubj}}}$ & won$_2$ & the$_3$  game$_4$ & PRP VBD & $\langle 2,1, \textrm{nsubj} \rangle$ \\
7 & \textproc{Shift} & won$_2$ the$_3$ & game$_4$ &  PRP VBD & $\langle 2,1, \textrm{nsubj} \rangle$ \\
8 & \textproc{Tag}$_{{\tiny \textrm{DT}}}$ & won$_2$ the$_3$ & game$_4$ &  PRP VBD DT & $\langle 2,1, \textrm{nsubj} \rangle$ \\
9 & \textproc{Shift} &  won$_2$ the$_3$ game$_4$ &  &  PRP VBD DT & $\langle 2,1, \textrm{nsubj} \rangle$ \\
10 & \textproc{Tag}$_{{\tiny \textrm{NN}}}$ &  won$_2$ the$_3$ game$_4$ &  &  PRP VBD DT NN & $\langle 2,1, \textrm{nsubj} \rangle$ \\
11 & \textproc{Left} & won$_2$ game$_4$ & & PRP  VBD  DT  NN & $\langle 2,1, \textrm{nsubj} \rangle$ $\langle 4,3, \bot \rangle$ \\
12 & \textproc{Label}$_{{\tiny \textrm{det}}}$ & won$_2$ game$_4$ & & PRP  VBD  DT  NN & $\langle 2,1, \textrm{nsubj} \rangle$ $\langle 4,3, \textrm{det} \rangle$ \\
13 & \textproc{Right} & won$_2$ & & PRP VBD DT NN & $\langle 2,1, \textrm{nsubj} \rangle$ $\langle 4,3, \textrm{det} \rangle$ $\langle 2,4, \bot \rangle$ \\
14 & \textproc{Label}$_{{\tiny \textrm{dobj}}}$ & won$_2$ & & PRP VBD DT NN & $\langle 2,1, \textrm{nsubj} \rangle$ $\langle 4,3, \textrm{det} \rangle$ $\langle 2,4, \textrm{dobj} \rangle$ \\
\hline
\end{tabular}

\caption{The process of joint POS tagging and dependency parsing for the example in Figure \ref{fig:example}.} \label{table:process}
\end{table*}

\subsection{Modeling}
Given a sentence $\mathbf{x}$ with $N$ words, tag sequence $\mathbf{t}$ and dependency tree $\mathbf{d}$ corresponds to a unique sequence of action-configuration pairs $\{ \langle c_{i}, a_{i} \rangle \}_{i=1}^{4N-2}$, as shown in Table \ref{table:process} \footnote{We follow Chen and Manning \shortcite{Chen:14} to map a parse to a unique sequence of action-configuration pairs by using the ``shortest stack'' strategy.}.
Note that the number of \textproc{Shift} actions is $N$, \textproc{Left} or \textproc{Right} is $N-1$, \textproc{Tag}$_t$ is $N$, and \textproc{Label}$_l$ is $N-1$,
where \textproc{Shift} and \textproc{Tag}$_t$ have the same number as words, \textproc{Left}/\textproc{Right} and \textproc{Label} have the same number as dependency arcs.

As a result, the probabilistic model for transition-based joint  POS tagging and dependency parsing is defined as
\begin{eqnarray}
 P(\mathbf{t}, \mathbf{d}|\mathbf{x}; \bm{\theta}) =\prod_{i=1}^{4N-2}P(a_i|c_{i-1}; \bm{\theta})\times P(c_i|c_{i-1},a_i)
\end{eqnarray}
%Note that $P(c_i|c_{i-1}, a_i) = 1$ if and only if $a_i$ is a legal transition between $c_{i-1}$ and $c_i$. Otherwise, $P(c_i|c_{i-1}, a_i) = 0$.
Therefore, we only need to focus on the action probability conditioned on the previous configuration.

In our transition system, there are three types of conflicts:
\begin{enumerate}
\item {\em Tag conflict} among all possible POS tags $\{$\textproc{Tag}$_t | t \in \mathcal{T} \}$,
\item {\em Shift/reduce conflict} between \textproc{Shift}, \textproc{Left}, and \textproc{Right}. For example, at step 5 in Table \ref{table:process}, both \textproc{Shift} and \textproc{Left} can be applied,
\item {\em Label conflict} among all possible syntactic labels $\{$\textproc{Label}$_l | l \in \mathcal{L} \}$.
\end{enumerate}

To resolve these conflicts, we develop three corresponding neural network based classifiers. Note that the separation of structural actions from tagging and labeling actions results in three small classifiers with fewer classes (i.e., $|\mathcal{T}|$  classes for the tag classifier, 3 for the shift/reduce classifier, and $|\mathcal{L}|$ for the label classifier) rather than one big classifier with much more classes (i.e., $|\mathcal{T}|+2|\mathcal{L}|$).

\subsubsection{Basic Features}
 We use $\bm{x}_n$ to denote the vector representation of the $n$-th word $x_n$. In our experiments, we follow Kiperwasser and Goldberg \shortcite{Kiperwasser:16} to learn $\bm{x}_n$ using bidirectional LSTM whose inputs are concatenations of randomly initialized word embeddings with additional pre-trained embeddings as well as character-based representations \cite{santos:2014,Ballesteros:15}. We use $\bm{t}_n$ to denote the vector representation of the $n$-th POS tag $t_n$, which can be learned using a unidirectional LSTM based on randomly initialized tag embeddings. Note that the bidirectional LSTM feature representations for words are computed before joint POS tagging and dependency parsing while the unidirectional LSTM feature representations for tags are calculated during the search on the fly.

\subsubsection{Tag Classification}
Resolving the tag conflict is a $|\mathcal{T}|$-class classification problem.
Instead of using conventional feature templates that are highly sparse and inevitably incomplete, we leverage a neural network based classifier. To determine the POS tag of the last word added to the stack, which is represented as $x_{S_{0}}$, the input layer consists of the following representations:
\begin{enumerate}
\item $\bm{x}_{S_1}$: the word representation of the second item in the stack,
\item $\bm{t}_{S_1}$: the tag representation of the second item in the stack,
\item $\bm{x}_{B_{-2}}$: the word representation of the second last item removed from the buffer,
\item $\bm{t}_{B_{-2}}$: the tag representation of the second last item removed from the buffer,
\item $\bm{x}_{S_{0}}$: the word representation of the first item in the stack,
\item $\bm{x}_{B_0}$: the word representation of the first item in the buffer.
\end{enumerate}
where, $\bm{x}_{B_{-2}}$, $\bm{x}_{S_{0}}$, $\bm{x}_{B_0}$ are window-based features that have been widely adopted in previous work \cite{huang+:2015} and
%todo: pre-tag
$\bm{t}_{B_{-2}}$ models the previous tag which has been widely used implicitly by markov assumption in CRF models.
Note that  $x_{B_{-2}}$, $x_{S_{0}}$, $x_{B_{0}}$ are sequential words and $x_{S_1}$ is not necessarily identical to $x_{B_{-2}}$ due to the \textproc{Right} action.
%For example, at step 8 in Figure \ref{table:process}, $x_{S_0}$ is ``the'', $x_{S_1}$ is ``won'', $x_{B_{-2}}$ is ``won'', $x_{S_0}$ is ``the'', $x_{B_0}$ is ``game''.

We expect that these representations can provide useful contextual information for resolving the tagging ambiguity. Note that the tagging classifier is capable of exploiting syntactic information encoded in $\bm{x}_{S_1}$ and $\bm{t}_{S_1}$.

\begin{figure}
  \centering
  \subfigure[Tag Classification.]{
    \begin{minipage}[b]{0.38\textwidth}
      \includegraphics[width=\textwidth]{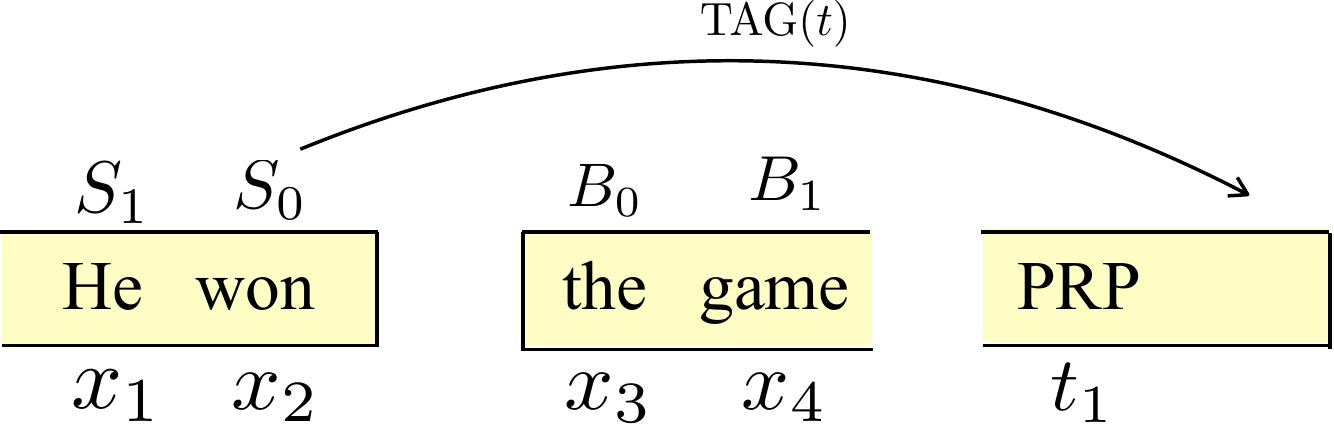}
    \end{minipage}
    \label{fig:tag}
  }

  \subfigure[ Shift/Reduce Classification.]{
    \begin{minipage}[b]{0.38\textwidth}
      \includegraphics[width=\textwidth]{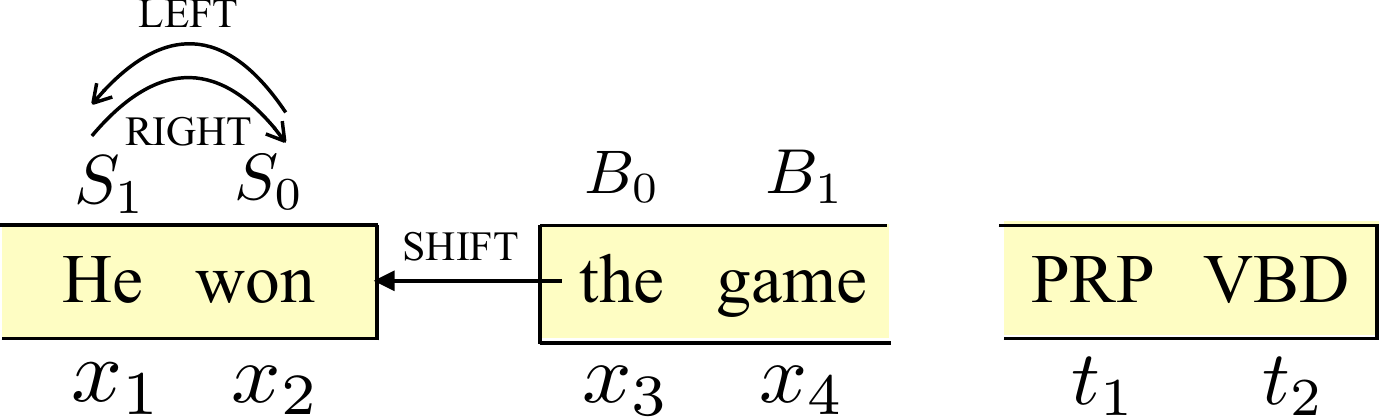}
    \end{minipage}
    \label{fig:shift}
  }

  \subfigure[Label Classification.]{
    \begin{minipage}[b]{0.38\textwidth}
      \includegraphics[width=\textwidth]{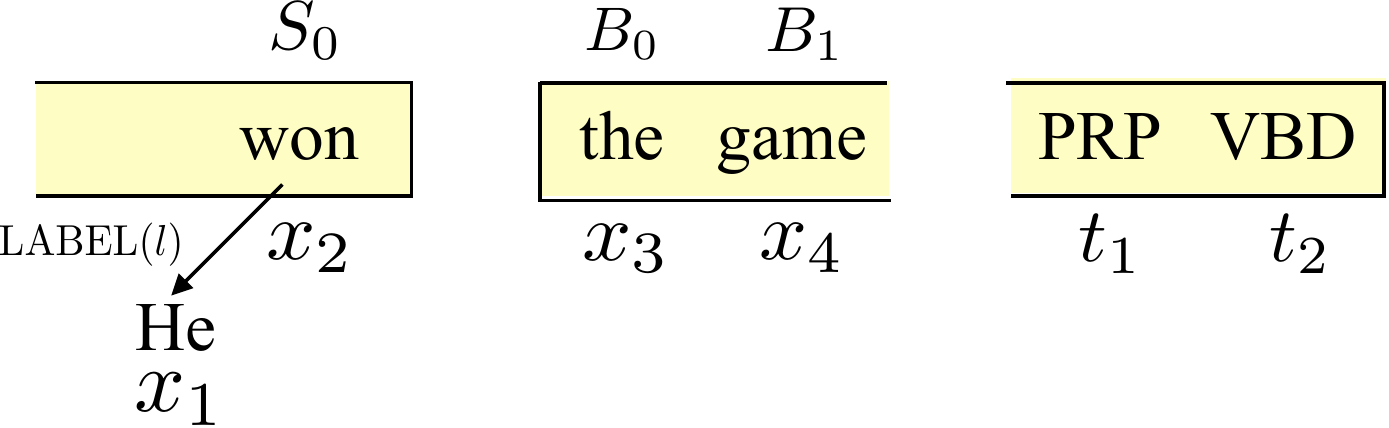}
    \end{minipage}
    \label{fig:label}
  }
  \caption{Examples to illustrate the three classifiers.}
\end{figure}

As shown in Figure \ref{fig:tag}, the hidden layer is calculated as
\begin{eqnarray}
\mathbf{h}^{S_0}_{\mathrm{tag}} = \mathbf{W}^{(1)}_{\mathrm{tag}}[\bm{x}_{S_1};\bm{t}_{S_1};\bm{x}_{B_{-2}};\bm{t}_{B_{-2}};\bm{x}_{S_{0}};\bm{x}_{B_0}] \label{eq:hidden_tag}
\end{eqnarray}

Then, the probability for tagging $x_{S_0}$ as $t$ is computed at the softmax layer:
\begin{eqnarray}
P_{\mathrm{tag}}(a |c; \bm{\theta}) = \mathrm{softmax}\Big(\mathbf{W}^{(2)}_{\mathrm{tag}}\mathbf{h}^{S_0}_{\mathrm{tag}} \Big)
\end{eqnarray}
where $a \in \{ \textrm{\textproc{Tag}}_t | t \in \mathcal{T} \}$.

\subsubsection{Shift/Reduce Classification}
Resolving the shift/reduce conflict is a $3$-class classification problem. As shown in Figure \ref{fig:shift}, we also use a neural classifier, in which the hidden layer is given by:
\footnote{Although it is possible to use hidden states in the tag classifier (e.g., $\mathbf{h}^{S_0}_{\mathrm{tag}}$) to replace tag representations $\bm{t}_{S_0}$ as suggested by Zhang and Weiss \shortcite{Zhang:16}, we find that it results in degenerate tagging and parsing results as compared with Eq. (\ref{eq:hidden_parse}).}
\begin{eqnarray}
\mathbf{h}_{\mathrm{parse}} = \mathbf{W}^{(1)}_{\mathrm{parse}}[\bm{x}_{S_2};\bm{t}_{S_2};\bm{x}_{S_1};\bm{t}_{S_1};\bm{x}_{S_0};\bm{t}_{S_0};\bm{x}_{B_0}] \label{eq:hidden_parse}
\end{eqnarray}
where $S_2$ denotes the third item in the stack. Note that the shift/reduce classifier is capable of exploiting lexical information encoded in $\bm{t}_{S_2}$, $\bm{t}_{S_1}$, and $\bm{t}_{S_0}$.

Therefore, the shift/reduce classification probability is computed as
\begin{eqnarray}
P_{\mathrm{parse}}(a|c; \bm{\theta})= \mathrm{softmax}\Big(\mathbf{W}^{(2)}_{\mathrm{parse}}\mathbf{h}_{\mathrm{parse}} \Big)
\end{eqnarray}
where $a \in \{ \textrm{\textproc{Shift}}, \textrm{\textproc{Left}}, \textrm{\textproc{Right}}  \}$.

\subsubsection{Label Classification}
Resolving the label conflict is a $|\mathcal{L}|$-class classification problem. As shown in Figure \ref{fig:label}, the corresponding neural classifier takes the word and tag representations of the first two items in the stack as input:
\begin{eqnarray}
\mathbf{h}_{\mathrm{label}} = \mathbf{W}^{(1)}_{\mathrm{label}}[\bm{x}_{S_1};\bm{t}_{S_1};\bm{x}_{S_0};\bm{t}_{S_0}] \label{eq:hidden_label}
\end{eqnarray}
Clearly, labeling a dependency arc also depends on tag representations $\bm{t}_{S_1}$ and $\bm{t}_{S_0}$.

The label classification probability is computed as
\begin{eqnarray}
P_{\mathrm{label}}(a|c; \bm{\theta})=\mathrm{softmax}\Big( \mathbf{W}^{(2)}_{\mathrm{label}} \mathbf{h}_{\mathrm{label}} \Big)
\end{eqnarray}
where $a \in \{ \textrm{\textproc{Label}}_l | l \in \mathcal{L} \}$.

\subsection{Training and Parsing}
Given a set of training examples $\{ \langle \mathbf{x}^{(k)}, \mathbf{t}^{(k)}, \mathbf{d}^{(k)} \rangle \}_{k=1}^{K}$, the training objective is to minimize the cross-entropy loss plus a $\ell_2$-regularization term:
\begin{eqnarray}
\hat{\bm{\theta}} = \argmin_{\bm{\theta}}\Big\{ -\sum_{k=1}^{K}\log P(\mathbf{t}^{(k)}, \mathbf{d}^{(k)}|\mathbf{x}^{(k)}; \bm{\theta}) + \frac{\lambda}{2} ||\bm{\theta}||^2 \Big\}
\end{eqnarray}
%We use mini-batched Adam \cite{Kingma:15} to optimize the model parameters.

In parsing, we follow Chen and Manning \shortcite{Chen:14} to perform greedy decoding. The most probable tag sequence and dependency tree corresponds to a sequence of action-configuration pairs with the highest probability: $\{ \langle \hat{c}_i, \hat{a}_i \rangle \}_{i=1}^{4N-2}$, where
\begin{eqnarray}
\hat{a}_i = \argmax_{a} P(a|\hat{c}_{i-1}; \hat{\bm{\theta}})
\end{eqnarray}
and $\hat{c}_i$ is obtained by applying $\hat{a}_i$ to $\hat{c}_{i-1}$.

\begin{table*}[!t]
\centering
\begin{tabular}{l|c|c|c|c|c|c}
\hline
\multirow{2}{*}{Method}  & \multicolumn{3}{c}{PTB} & \multicolumn{3}{|c}{CTB5} \\
\cline{2-7}
& POS & UAS & LAS & POS & UAS & LAS \\
\hline \hline
\multicolumn{7}{c}{\em Joint models} \\ \hline 
%Li et al. \shortcite{Li:11} & -- & -- & -- & 93.08 & 80.74 & --  \\
Hatori et al. \shortcite{Hatori:11} & -- & -- & -- & 93.94 & 81.33 & -- \\
Bohnet and Nivre \shortcite{Bohnet:12} & 97.42 & 93.67 & 92.68 & 93.24 & 81.42 & 77.91 \\
Zhang and Weiss \shortcite{Zhang:16} & -- & 93.43 & 91.41 & -- & -- & -- \\
{\em this work (Joint)} & \bf{97.54} & \bf{94.18} & \bf{92.26} & \bf{95.58} & \bf{83.99} & \bf{81.39} \\ \hline 
\multicolumn{7}{c}{\em{Pipeline models}} \\ \hline 
Dyer et al. \shortcite{Dyer:15} & -- & 93.10 & 90.90 & 100 & 87.20 & 85.70 \\
Kiperwasser and Goldberg \shortcite{Kiperwasser:16} & -- & 93.90 & 91.90 & 100 & 87.60 & 86.10 \\
Andor et al. \shortcite{Andor:16} & -- & 94.61 & 92.79 & -- & -- & --  \\
Chen et al. \shortcite{cheng+:2016} & -- & 94.10 & 91.49 & 100 & 88.10 & 85.70  \\ \hline
%Dozen and Manning \shortcite{dozat+manning:2017} & -- & 95.44 & 93.76 & -- & 90.03 & 86.42\\ \hline
{\em this work (auto POS)} & 97.45 & 93.74 & 91.32 & 95.06 & 82.68 & 79.93 \\ 
{\em this work (gold POS)} & 100 & 94.73 & 93.53 & 100 & 88.75 & 87.53 \\ 
\hline 
\end{tabular}
\caption{Final results on the datasets of PTB and CTB5.1, where the tagging accuracy being 100\% denotes gold-standard POS tags are employed. We include the results of state-of-the-art previous transition-based parsers as well. In particular, Andor et al. (2016) use beam search in decoding and Bohnet and Nivre (2012) use a different method to produce dependency trees. } \label{table:comparison_j}
\end{table*}

\begin{table*}[!t]
\centering
\begin{tabular}{l|c|c|c|c|c|c|c|c}
\hline
Method  & de & en & es & fr & it & pt & sv & AVG \\
\hline 
Ballesteros et al. \shortcite{Ballesteros:15}  & 73.0 & 77.9 & 77.8 & 78.0 & 84.2 & 80.4 & 74.5 &   78.0\\ 
Zhang and Weiss \shortcite{Zhang:16}  & 74.2 & 80.7 & 80.7 & 80.0 & 85.8 & 80.4 & 77.5 & 79.9  \\ 
\hline
{\em this work (Pipeline)} & 74.6 & 80.6 &80.6 & 78.9 & 84.9 & 81.6 & 77.6 & 79.8 \\
{\em this work (Joint)} & \bf{77.1} & \bf{82.5} & \bf{82.5} & \bf{81.2} & \bf{87.0} & \bf{83.1} & \bf{80.4} & \bf{82.0}  \\
\hline 

\end{tabular}
\caption{Final dependency parsing results (LAS) on the UD dataset. } \label{table:comparison_ud}
\end{table*}

\section{Experiments}

\subsection{Setup}
\subsubsection{Datasets and Evaluation}
We evaluate our approach on three datasets:  the English Penn Treebank (PTB) with annotated phrase-structure trees of English, the Chinese Penn Treebank (CTB) version 5.1 with annotated phrase-structure trees of Chinese,
and the Universal Dependency Treebank (UD) version 1.2 \footnote{http://universaldependencies.org} with annotated dependency trees across a number of natural languages.

We use the standard splitting method to divide the PTB dataset into training, development and test sections,
and convert the phrase-structure trees into dependency trees by the Stanford dependency converter v3.3.0 \cite{deMarneffe:06}.
For the CTB5.1 dataset, we follow previous work \cite{Hatori:11,Bohnet:12}  to split the dataset into training, development and test sections ,
and use the Penn2Malt tool with the head-finding rule of \cite{Zhang:08} to convert the phrase-structure trees into dependencies.
For the UD dataset, we follow Ammar et al. \shortcite{Ammar:16}, using the same subset of seven languages including German (de), English (en), Spanish (es), French (fr), Italian (it), Portuguese (pt) and Swedish (sv) and using the same data splitting method.

For POS tagging, we use the standard tagging accuracy (POS) based on words as the major evaluation metric.
For dependency parsing, we use two metrics, namely unlabeled attachment score (UAS) and labeled attachment score (LAS),
where UAS denotes the ratio of the correctly-headed words with respect to the total words,
which considers only the head of a word,
and LAS takes into account the dependency label as well,
and is emlpoyed as the major metric to evaluate dependency parsing.

\subsubsection{Hyper-parameters and Training Details}
We tune all hyper-parameters in our models according the development results.
Concretely, the dimension sizes of word, tag and character embeddings are 150, 50 and 50, respectively.
We use the same pre-trained word embeddings for PTB and CTB5.1 as Chris et al. \shortcite{Dyer:15} \footnote{We thank the authors very much for sharing their data with us. },
and do not use any pre-trained embeddings for UD,
and the dimension size of the hidden states in neural classifiers is 300.

We exploit the Adam optimizer \cite{Kingma:15} to update model parameters during training,
setting the hyper-parameters $\beta_1$ and $\beta_2$ both to 0.9.
Gradient clipping \cite{Pascanu:13} by a max norm 5.0 is used to avoid gradient exploding.
To avoid overfitting, we use $\ell_2$-regularization by a parameter $10^{-8}$
as well as the dropout technique \cite{Srivastava:14} with a drop rate of $0.25$.
Since the arc-standard algorithm can only handle the projective trees,
we apply a projectivization step to the training sets of the UD dataset.

%In particular, we adopt a pre-training method \cite{Andor:16,Wiseman:16} to better initialize the model parameters of the joint model: the tagging sub-model is first trained for 30 iterations and the parsing sub-model is then trained for 40 iterations before training the joint model.

\subsection{Main Results}

Table \ref{table:comparison_j} shows the final results of our models on PTB and CTB5.1.
We include the pipeline performances as well.
Our joint model brings significant improvements on both POS tagging (POS) and dependency parsing (LAS) compared with the pipeline model (the p-value is below $10^{-5}$ using pairwise t-test).
In addition, we compare our joint model with the baseline parsing model using gold-standard POS tags,
which can be treated as the oracle performances of our joint model.
Although the joint model gives improved performances over the pipeline model, 
it still has large spaces to reach the oracle performances,
which demonstrates the effectiveness of POS tags in dependency parsing.

We compare our model with previous work as well.
On the one hand,
we compare our joint model with previous joint models.
As shown in Table \ref{table:comparison_j}, 
our neural joint model shows the highest results for both PTB and CTB5.1, 
obtaining much higher performances in dependency parsing,
which demonstrates the effectiveness of the neural features. 
On the other hand,
we compare our baseline model with state-of-the-art transition-based dependency parsing models.
%\footnote{Recently top-performing dependency parsing results are achieved by graph-based models \cite{dozat+manning:2017}. We do not compare with them because they suffer from the low efficiency problem.}
Typically, the PTB results are reported by using auto POS tags and the CTB5.1 results are reported by using gold-standard POS tags, respectively.
Our baseline model produces strong enough results for both PTB and CTB5.1.

Table \ref{table:comparison_ud} shows the final results on the UD dataset.
Joint models also achieves significantly better results in comparison with the pipeline models (p-value below $10^{-5}$),
which is similar to our finding on CTB5.1.
Besides, our joint model achieves the best-reported results among the transition-based models,
even by using a greedy manner for decoding,
which can be attributed to the effective exploration of the interaction between the tagging and parsing in our joint model,
while no previous work has studied it under the neural setting to our knowledge.
The work of Zhang and Weiss \shortcite{Zhang:16} resembles our work most,  
which improve a feed-forward dependency parser by using POS tags in a pipeline way by stack-propagation.
While our joint model benefits from the use of LSTM, 
and in addition, we find that directly using the resulting tags rather than the penultimate hidden representations of a tag classifier leads to better results.

%For English, Bohnet and Nivre \shortcite{Bohnet:12} use the head rules of Yamada and Matsumoto \shortcite{Yamada:03} while Zhang and Weiss \shortcite{Zhang:16} and our approach use the Stanford head rules \cite{deMarneffe:06}. For Chinese, all methods use the head rules of Zhang and Clark \shortcite{Zhang:08}.

%Our approach generally outperforms all previous methods, either using conventional feature templates \cite{Li:11,Hatori:11,Bohnet:12} or using neural networks \cite{Zhang:16}. Note that Bohnet and Nivre \shortcite{Bohnet:12}'s results on the English Penn Treebank are not strictly comparable to our approach due to the difference in head rules.

\subsection{Discussion}
%We conduct analysis on the CTB5.1 dataset to illustrate the effectiveness of the joint models.

\begin{table}[!t]
\centering
\begin{tabular}{|c|c||c|c|c|}
\hline
\multicolumn{2}{|c||}{Interaction} & \multirow{2}{*}{POS} & \multirow{2}{*}{UAS} & \multirow{2}{*}{LAS} \\
\cline{1-2}
tag $\rightarrow$ parse & tag $\leftarrow$ parse &  &  &  \\
\hline\hline
$\times$ & $\times$ & 95.19 & 83.38 & 80.66  \\
$\times$ & $\surd$ & 95.25 & 83.56 & 80.82 \\
$\surd$ & $\times$ & 95.50 & 84.10 & 81.59 \\
$\surd$ & $\surd$ & \bf 95.63 & \bf 84.20 & \bf 81.76 \\
\hline
\end{tabular}
\caption{Interaction between POS tagging and dependency parsing. ``tag $\rightarrow$ parse'' denotes that parsing leverages lexical information and ``tag $\leftarrow$ parse'' denotes that tagging exploits syntactic information. The interactions can be disabled as shown in Eq. (\ref{eq:notag_p})-(\ref{eq:notag_t}). The tagging and parsing results are evaluated on the Chinese Penn Treebank development set.} \label{table:interaction}
\end{table}

%\subsubsection{Interaction between Tagging and Parsing}
To investigate the effect of POS tagging on dependency parsing,
we conduct analysis on the CTB5.1 dataset to illustrate the effectiveness of the joint model.
Here we examine in detail to see the benefits from the interaction between tagging and parsing in our joint model.
First, we can remove the tag representations from parsing in Eq. (\ref{eq:hidden_parse}) and Eq. (\ref{eq:hidden_label}):
\begin{eqnarray}
\tilde{\mathbf{h}}_{\mathrm{parse}} &=& \tilde{\mathbf{W}}^{(1)}_{\mathrm{parse}}[\bm{x}_{S_2};\bm{x}_{S_1}; \bm{x}_{S_0}; \bm{x}_{B_0}] \label{eq:notag_p} \\
\tilde{\mathbf{h}}_{\mathrm{label}} &=& \tilde{\mathbf{W}}^{(1)}_{\mathrm{label}}[\bm{x}_{S_1}; \bm{x}_{S_0}] \label{eq:notag_l}
\end{eqnarray}

Similarly, we can also remove the syntactic information from tagging in Eq. (\ref{eq:hidden_tag}) to investigate the effect of dependency parsing on POS tagging:
\begin{eqnarray}
\tilde{\mathbf{h}}_{\mathrm{tag}}^{S_0} = \tilde{\mathbf{W}}^{(1)}_{\mathrm{tag}}[\bm{x}_{B_{-2}}; \bm{t}_{B_{-2}};\bm{x}_{S_0};\bm{x}_{B_0}] \label{eq:notag_t}
\end{eqnarray}

Table \ref{table:interaction} gives the tagging and parsing results on the CTB 5.1 development set. We observe that disabling the interactions between tagging and parsing significantly deteriorates both tagging and parsing quality.

An interesting finding is that providing lexical information to parsing (``tag $\rightarrow$ parse'') leads to more benefits than providing syntactic information to tagging (``tag $\leftarrow$ parse''). This is because tagging ambiguity is mostly local while dependency parsing heavily depends on POS tags to predict syntactic structures.

Note that enabling ``tag $\rightarrow$ parse'' only also improves the tagging accuracy itself. One possible reason is that tagging and parsing is still connected via the sharing of word embeddings and bidirectional LSTM hidden states although the connection at hidden layer in classifiers is explicitly disabled.

%Following Zhang and Weiss \shortcite{Zhang:16}, we also observe a significant reduction in cascaded POS tagging errors. On CTB5 test set, the gain in LAS is 10.8\% (43.3\% vs. 54.1\%) on tokens where the independently trained tagger makes a mistake.

\section{Related Work}

Our work is closely related to two lines of research: (1) joint POS tagging and dependency parsing using feature templates, and (2) neural dependency parsing.

\subsection{Joint Modeling with Feature Templates}
Most previous endeavors on  joint POS tagging and dependency parsing have focused on developing linear models with feature templates \cite{Li:11,Hatori:11,Bohnet:12}. They introduce transition systems that can perform POS tagging and dependency parsing in a joint search space.

 Our transition system differs from previous work in the separation of structural, tagging, and labeling actions. This results in three small classifiers with fewer classes (i.e., $|\mathcal{T}|$  classes for the tag classifier, 3 for the shift/reduce classifier, and $|\mathcal{L}|$ for the label classifier) rather than one big classifier with much more classes (i.e., $|\mathcal{T}|+2|\mathcal{L}|$).

More importantly, we use continuous representations instead of discrete indicator features to build the classifiers. As indicated by Chen and Manning \shortcite{Chen:14}, lexicalized indicator features crucial for improving parsing accuracy are highly sparse and often incomplete. Alternatively, we resort to neural networks to learn representations from data to circumvent the sparsity and incompleteness problems. Another benefit of using neural networks is that there is no need to compose individual features to obtain more complex features like conventional discriminative dependency parsing \cite{Dyer:15}.

\subsection{Neural POS Tagging and Dependency Parsing}

Our work is also inspired by recent advances in applying neural networks to POS tagging \cite{huang+:2015} and dependency parsing \cite{Chen:14,Dyer:15,Ballesteros:15,alberti+:2015,Ammar:16,Kiperwasser:16,Andor:16,wang+chang:2016,cheng+:2016,dozat+manning:2017}.

Among them, our work bears the most resemblance to \cite{Zhang:16}, which propose stack-propagation to integrate a tagging model into a neural parser. They propose a stacked pipeline of models and utilize POS tags as a regularizer of learned representations. While Zhang and Weiss \shortcite{Zhang:16} use the hidden layer of the tagger network as the input for the parser, we are interested in enabling tagging and parsing to benefit each other in a joint search space. As a result, the tagger is able to resolve long-distance tagging ambiguity by exploiting syntactic information. Meanwhile, the error propagation problem the parser faces can be alleviated due to the cascaded error reduction by joint modeling.

\section{Conclusion}
We have presented an approach to joint part-of-speech tagging and dependency parsing using transition-based neural networks. Based on a five-action transition system, we develop three classifiers to resolve structural, tagging, and labeling conflicts. As our approach allows lexicality and syntax to interact with each other in the joint search process, it improves over previous work on joint POS tagging and dependency parsing on three treebanks across a variety of natural languages.
Our code is released at \url{http://github.com/lineryang/joint-parser}.

%% The file named.bst is a bibliography style file for BibTeX 0.99c
\bibliographystyle{named}
\bibliography{ijcai17_yle}

\begin{thebibliography}{}

\bibitem[\protect\citeauthoryear{Alberti \bgroup \em et al.\egroup
  }{2015}]{alberti+:2015}
Chris Alberti, David Weiss, Greg Coppola, and Slav Petrov.
\newblock Improved transition-based parsing and tagging with neural networks.
\newblock In {\em Proceedings of EMNLP}, 2015.

\bibitem[\protect\citeauthoryear{Ammar \bgroup \em et al.\egroup
  }{2016}]{Ammar:16}
Waleed Ammar, George Mulcaire, Miguel Ballesteros, Chris Dyer, and Noah~A.
  Smith.
\newblock Many languages, one parser.
\newblock {\em TACL}, 2016.

\bibitem[\protect\citeauthoryear{Andor \bgroup \em et al.\egroup
  }{2016}]{Andor:16}
Daniel Andor, Chris Alberti, David Weiss, Aliasei Severyn, Alessandro Presta,
  Kuzman Ganchev, Slav Petrov, and Michael Collins.
\newblock Globally normalized transition-based neural networks.
\newblock In {\em Proceedings of ACL}, 2016.

\bibitem[\protect\citeauthoryear{Ballesteros \bgroup \em et al.\egroup
  }{2015}]{Ballesteros:15}
Miguel Ballesteros, Chris Dyer, and Noah Smith.
\newblock Improved transition-based parsing by modeling characters instead of
  words with lstms.
\newblock In {\em Proceedings of EMNLP}, 2015.

\bibitem[\protect\citeauthoryear{Bohnet and Nivre}{2012}]{Bohnet:12}
Bernd Bohnet and Joakim Nivre.
\newblock A transition-based system for joint part-of-speech tagging and
  labeled non-projective parsing.
\newblock In {\em Proceedings of EMNLP}, 2012.

\bibitem[\protect\citeauthoryear{Chen and Manning}{2014}]{Chen:14}
Danqi Chen and Christopher~D. Manning.
\newblock A fast and accurate dependency parser using neural networks.
\newblock In {\em Proceedings of EMNLP}, 2014.

\bibitem[\protect\citeauthoryear{Cheng \bgroup \em et al.\egroup
  }{2016}]{cheng+:2016}
Hao Cheng, Hao Fang, Xiaodong He, Jianfeng Gao, and Li~Deng.
\newblock Bi-directional attention with agreement for dependency parsing.
\newblock In {\em Proceedings of EMNLP}, 2016.

\bibitem[\protect\citeauthoryear{Collins}{2002}]{collins:2002}
Michael Collins.
\newblock Discriminative training methods for hidden markov models: Theory and
  experiments with perceptron algorithms.
\newblock In {\em Proceedings of EMNLP}, 2002.

\bibitem[\protect\citeauthoryear{de Marneffe \bgroup \em et al.\egroup
  }{2006}]{deMarneffe:06}
Marie-Catherine de~Marneffe, Bill MacCartney, and Christopher~D. Manning.
\newblock Generating typed dependency parses from phrase structure parses.
\newblock In {\em Proceedings of LREC}, 2006.

\bibitem[\protect\citeauthoryear{dos Santos and Zadrozny}{2014}]{santos:2014}
C{\'i}cero~Nogueira dos Santos and Bianca Zadrozny.
\newblock Learning character-level representations for part-of-speech tagging.
\newblock In {\em Proceedings of ICML}, 2014.

\bibitem[\protect\citeauthoryear{Dozat and Manning}{2017}]{dozat+manning:2017}
Timothy Dozat and Christopher~D. Manning.
\newblock Deep biaffine attention for neural dependency parsing.
\newblock In {\em Proceedings of ICLR}, 2017.

\bibitem[\protect\citeauthoryear{Dyer \bgroup \em et al.\egroup
  }{2015}]{Dyer:15}
Chris Dyer, Miguel Ballesteros, Wang Ling, Austin Mattews, and A.~Smith, Noah.
\newblock Transition-based depdnency parsing with stack long short-term memory.
\newblock In {\em Proceedings of ACL}, 2015.

\bibitem[\protect\citeauthoryear{Hatori \bgroup \em et al.\egroup
  }{2011}]{Hatori:11}
Jun Hatori, Takuya Matsuzaki, Yusuke Miyao, and Jun'ichi Tsujii.
\newblock Incremental joint pos tagging and dependency parsing in chinese.
\newblock In {\em Proceedings of IJCNLP}, 2011.

\bibitem[\protect\citeauthoryear{Huang \bgroup \em et al.\egroup
  }{2015}]{huang+:2015}
Zhiheng Huang, Wei Xu, and Kai Yu.
\newblock Bidirectional lstm-crf models for sequence tagging.
\newblock {\em arXiv}, 2015.

\bibitem[\protect\citeauthoryear{Kingma and Ba}{2015}]{Kingma:15}
Diederik~P. Kingma and Jimmy Ba.
\newblock Adam: A method for stochastic optimization.
\newblock In {\em Proceedings of ICLR}, 2015.

\bibitem[\protect\citeauthoryear{Kiperwasser and
  Goldberg}{2016}]{Kiperwasser:16}
Eliyahu Kiperwasser and Yoav Goldberg.
\newblock Simple and accurate dependency parsing using bidirectional lstm
  feature representations.
\newblock {\em TACL}, 2016.

\bibitem[\protect\citeauthoryear{Li \bgroup \em et al.\egroup }{2011}]{Li:11}
Zhenghua Li, Min Zhang, Wanxiang Che, Ting Liu, Wenliang Chen, and Haizhou Li.
\newblock Joint models for chinese pos tagging and dependency parsing.
\newblock In {\em Proceedings of EMNLP}, 2011.

\bibitem[\protect\citeauthoryear{Li \bgroup \em et al.\egroup }{2012}]{Li:12}
Zhenghua Li, Min Zhang, Wanxiang Che, and Ting Liu.
\newblock A separately passive-aggressive training algorithm for joint pos
  tagging and dependency parsing.
\newblock In {\em Proceedings of COLING}, 2012.

\bibitem[\protect\citeauthoryear{McDonald \bgroup \em et al.\egroup
  }{2005}]{McDonald:05}
Ryan McDonald, Koby Crammer, and Fernando Pereira.
\newblock Online large-margin training of dependency parsers.
\newblock In {\em Proceedings of ACL}, 2005.

\bibitem[\protect\citeauthoryear{Nivre \bgroup \em et al.\egroup
  }{2006}]{nivre+:2006}
Joakim Nivre, Johan Hall, Jens Nilsson, Gulsen Eryigit, and Svetoslav Marinov.
\newblock Labeled pseudo-projective dependency parsing with support vector
  machines.
\newblock In {\em Proceedings of CoNLL}, 2006.

\bibitem[\protect\citeauthoryear{Nivre}{2008}]{Nivre:08}
Joakim Nivre.
\newblock Algorithms for deterministic incremental depdendency parsing.
\newblock {\em Computational Linguistics}, 2008.

\bibitem[\protect\citeauthoryear{Pascanu \bgroup \em et al.\egroup
  }{2013}]{Pascanu:13}
R.~Pascanu, T.~Mikolov, and Y.~Bengio.
\newblock On the difficulty of training recurrent neural networks.
\newblock In {\em Proceedings of ICML}, 2013.

\bibitem[\protect\citeauthoryear{Srivastava \bgroup \em et al.\egroup
  }{2014}]{Srivastava:14}
Nitish Srivastava, Geoffrey Hinton, Alex Krizhevsky, Ilya Sutskever, and Ruslan
  Salakhtdinov.
\newblock Dropout: A simple way to prevent neural networks from overfitting.
\newblock {\em Journal of Machine Learning Research}, 2014.

\bibitem[\protect\citeauthoryear{Sun \bgroup \em et al.\egroup
  }{2013}]{sun+:2013}
Weiwei Sun, Xiaochang Peng, and Xiaojun Wan.
\newblock Capturing long-distance dependencies in sequence models: A case study
  of chinese part-of-speech tagging.
\newblock In {\em Proceedings of IJCNLP}, 2013.

\bibitem[\protect\citeauthoryear{Toutanova \bgroup \em et al.\egroup
  }{2003}]{toutanova+:2003}
Kristina Toutanova, Dan Klein, Christopher~D. Manning, and Yoram Singer.
\newblock Feature-rich part-of-speech tagging with a cyclic dependency network.
\newblock In {\em Proceedings of NAACL}, 2003.

\bibitem[\protect\citeauthoryear{Wang and Chang}{2016}]{wang+chang:2016}
Wenhui Wang and Baobao Chang.
\newblock Graph-based dependency parsing with bidirectional lstm.
\newblock In {\em Proceedings of ACL}, 2016.

\bibitem[\protect\citeauthoryear{Zhang and Clark}{2008}]{Zhang:08}
Yue Zhang and Stephen Clark.
\newblock A tale of two parsers: Investigating and combining graph-based and
  transition-based dependency parsing.
\newblock In {\em Proceedings of EMNLP}, 2008.

\bibitem[\protect\citeauthoryear{Zhang and Weiss}{2016}]{Zhang:16}
Yuan Zhang and David Weiss.
\newblock Stack-propagation: Improved representation learning for syntax.
\newblock In {\em Proceedings of ACL}, 2016.

\bibitem[\protect\citeauthoryear{Zhang \bgroup \em et al.\egroup
  }{2012}]{zhang+:2012}
Meishan Zhang, Wanxiang Che, Ting Liu, and Zhenghua Li.
\newblock Stacking heterogeneous joint models of {C}hinese {POS} tagging and
  dependency parsing.
\newblock In {\em Proceedings of COLING}, 2012.

\end{thebibliography}

\end{document}